\documentstyle[11pt]{article}  

\def\auxil{\Leftarrow}

\def\largelydominates{\succeq} 

\def\largelydominatedby{\preceq} 

\def\informedby{\leftarrow} 

\def\Pinf{P^{\infty}}
\def\astar{A^{\star}}
\def\hone{h_{1}}
\def\htwo{h_{2}}

\def\astartwo{\astar(P \informedby P'')}
\def\astaronetwo{\astar(P \informedby P' \informedby P'')}
\def\problem{P}
\newenvironment{equation*}{\begin{displaymath}}{\end{displaymath}}
\newtheorem{theorem}{Theorem}{{}}

\input{epsf}

\begin{document}


\addtocounter{page}{-1}   

\title{A New Result on the Complexity\\ of Heuristic Estimates for the
$\astar$ Algorithm\thanks{This paper was published in Artificial Intelligence, 55, 1 (May 1992), 129-143. 
This research was made possible by support
from Heuristicrats Research Inc.~and the National Aeronautics and
Space Administration.  The first two authors may be reached at
$\left\{{\rm othar,mayer}\right\}$@heuristicrat.com.  The third
author may be reached at mgv@cs.scarolina.edu.}}

\author{
\ \\
\parbox{2in}{
\centerline{Othar Hansson \ \ and \ \ Andrew Mayer}
\centerline{Computer Science Division}
\centerline{University of California}
\centerline{Berkeley, CA  94720}
}
\hskip 0.75in
\parbox{2in}{
\centerline{Marco Valtorta}
\centerline{Department of Computer Science}
\centerline{University of South Carolina}
\centerline{Columbia, SC 29208}
}
}

\date{}

\maketitle
\thispagestyle{empty}

\centerline{\bf Abstract}

Relaxed models are abstract problem descriptions generated by
ignoring constraints that are present in base-level problems. They
play an important role in planning and search algorithms, as it has
been shown that the length of an optimal solution to a relaxed model
yields a monotone heuristic for an $\astar$ search of a base-level
problem.  Optimal solutions to a relaxed model may be computed
algorithmically or by search in a further relaxed model, leading to a
search that explores a hierarchy of relaxed models.

In this paper, we review the traditional definition of problem
relaxation and show that searching in the abstraction hierarchy
created by problem relaxation will not reduce the computational
effort required to find optimal solutions to the base-level problem,
unless the relaxed problem found in the hierarchy can be transformed
by some optimization (e.g., subproblem factoring).  Specifically, we
prove that any $\astar$ search of the base-level using a heuristic
$\htwo$ will {\em largely dominate} an $\astar$ search of the
base-level using a heuristic $\hone$, if $\hone$ must be computed by
an $\astar$ search of the relaxed model {\em using} $\htwo$.

\newpage

\section{Introduction}

We extend a result by Valtorta~\cite{valtorta84} on the complexity of
heuristic estimates for the $\astar$ algorithm in a way that makes it
applicable to some problems that have arisen in recent work in
planning and search~\cite{yung,mostow}.  We start by giving an
informal description of the new result that contrasts it with the
original one.  Definitions and proofs will be given in the body of
the paper.

Both our paper and~\cite{valtorta84} present complexity results in
the state-space model of problem-solving.  Both use the number of
node expansions performed by the algorithm as a measure of the cost
of solving a (state-space) problem using a (heuristic or blind)
search algorithm.  (By a ``blind'' search algorithm we mean one that
uses no cost estimates or uninformative cost estimates.)  Both
consider the cost of solving a problem as the total cost of computing
the heuristic plus the cost of using $\astar$ with that heuristic.
In both cases, the heuristics are viewed as solutions of relaxed
subproblems~\cite{pearl}, also known as auxiliary
problems~\cite{guida}\footnote{The relaxed subproblems
of~\cite{pearl} and the auxiliary problems of~\cite{guida} are
special cases of the abstracted problems of~\cite{mostow,prieditis}.
We do not claim that our results apply for the more general
definition of~\cite{mostow,prieditis}, except in the special cases
where that definition coincides with the older ones.}.

Valtorta's 1984 work was motivated by the attempt to spur research on
automating the efficient computation of heuristics by showing that,
with (certain classes of) inefficient heuristics, $\astar$ was no
better than brute-force search (Dijkstra's algorithm).  This paper is
motivated by the need to determine how best to use (in a sense to be
explained later) a set of available heuristics.

Let $P$, $P'$, $P''$ be three problems such that $P'' \auxil P'
\auxil P$, where the relation $\auxil$ is the auxiliarity relation.
Informally, this notation indicates that $P$ is the base problem,
$P'$ is a relaxed subproblem of $P$, and $P''$ is a relaxed
subproblem of $P'$.  A natural question is that of how to choose a
relaxed subproblem (of which there are many) so that the cost of
computing the heuristic (by solving the relaxed subproblem) plus the
cost of solving the base problem (by $\astar$ using the heuristic) is
minimum.  This is the question considered in~\cite{valtorta80}
and~\cite{valtorta84}, whose key result we summarize after the
introduction of necessary notation.

Let $A \largelydominates B$ denote the fact that algorithm $A$
``largely dominates'' algorithm $B$, i.e., in a certain technical
sense (to be explained later), $A$ is at least as efficient as $B$.

Let $\astar(P \informedby Alg(P'))$ denote the $\astar$ algorithm,
applied to the problem $P$, informed by heuristics derived from the
relaxed model $P'$, where instances of $P'$ are solved by an
algorithm.  If instances of $P'$ are solved by an $\astar$ search,
using $P''$, we use the notation $\astar(P \informedby \astar(P'
\informedby Alg(P'')))$.  The pattern of this notation allows us to
abbreviate this as $\astar(P \informedby P' \informedby P'')$, i.e.,
as a hierarchy of relaxed models, where the last relaxed model is
solved algorithmically.  (See Figure~\ref{twoalgs}.)

\begin{figure}[htbp]
\begin{center}
\epsfysize=2.5in\leavevmode\epsfbox{valtorta.idraw}
\end{center}
\caption{Relaxation hierarchies used in Algorithms $\astaronetwo$ and $\astartwo$}
\label{twoalgs}
\end{figure}

Let $\Pinf$ denote an element of the class of most-relaxed models,
i.e., a state-space graph in which there is a path of zero cost
between every pair of states.  $\astar(P \informedby \Pinf)$ is said
to be a blind or brute-force algorithm, as it uses a completely
uninformative heuristic.  Since Dijkstra's algorithm is optimal in a
large class of blind search algorithms, as shown
in~\cite{valtorta84}, by $\astar(P \informedby \Pinf)$ we mean
$\astar$ with the tie-breaking rule (for node expansions) that makes
it equivalent to Dijkstra's algorithm.

The original result~\cite{valtorta84} is:

\begin{equation}\label{neweqn1}
\forall P'\auxil P \;\;\; \astar(P \informedby P' \informedby \Pinf)
\largelydominatedby \astar(P \informedby \Pinf).
\end{equation}

The new result (which summarizes both Theorems 1 and 2) is:

\begin{equation}\label{neweqn2}
\forall P''\auxil P' \auxil P \;\;\; \astar(P \informedby P'
\informedby P'') \largelydominatedby \astar(P \informedby P''),
\end{equation}

and as a simple corollary,

\begin{equation}\label{neweqn3}
\forall P^n\auxil P^{n-1} \auxil \ldots \auxil P' \auxil P \;\;\;
\astar(P \informedby P' \informedby \dots \informedby P^{n-1}
\informedby P^n) \largelydominatedby \astar(P \informedby P^n).
\end{equation}

In words, while Valtorta showed that there is no efficient way of
solving a problem with $\astar$ when the cost of computing the
heuristic using a blind method is counted, we now show that the most
efficient way of using a hierarchy of relaxed models, culminating in
a heuristic that can be computed algorithmically, is to ignore the
hierarchy and use that heuristic directly.  Generating a relaxed
model is useful only when an efficient algorithm accompanies it.

The rest of the paper is organized as follows: In the second section
we provide background material on state-space search, the $\astar$
algorithm, and relaxed models.  The aim of the third section is to
motivate the need for our new result, by describing an instance from
the recent literature to which it can be usefully applied.  In the
fourth section, we prove our main result (corresponding to
(\ref{neweqn2}) above).  The fifth section, which concludes the
paper, is devoted to an interpretation of the main result, including
some remarks on the measure of cost (``largely dominates'') used to
evaluate the comparative efficiency of state-space search algorithms.

\section{Background}

\subsection{State-Space Search}

The state-space approach to problem-solving considers a problem as a
quadruple, $(S, O \subset S \times S, I \in S, G \subset S)$.  $S$ is
the set of possible {\em states} of the problem.  $O$ is the set of
{\em operators}, or transitions from state to state.  $I$ is the one
{\em initial state} of a problem instance, and $G$ is the set of {\em
goal states}.  Search problems can be represented as a state-space
graph, where the states are nodes, and the operators are directed,
weighted arcs between nodes (the weight associated with each
operator, $O_i$, is the cost of applying it, $C(O_i)$).  The problem
consists of determining a sequence of operators, $O_1, O_2, ..., O_n$
that, when applied to $I$, yields a state in $G$.  Such a sequence is
called a {\em solution path} (or {\em solution}), with {\em length}
$n$ and {\em cost} $\sum_{i=1}^{n} C(O_i)$.  A solution with minimum
cost is called {\em optimal}.

\subsection{$\astar$ Algorithm}

Solutions to a given problem may be found by brute force search over
the state-space.  However, as the sizes of the state-spaces of most
problems are prohibitively large, the only hope of finding an optimal
solution in reasonable time is to use an intelligent method of
guiding a search through the state-space.  Typically, such methods
take the form of branch-and-bound, wherein partial solutions
(equivalently, classes of solutions) are enumerated (``branch''), and
perhaps eliminated from future consideration by an estimate of
solution cost (``bound'').  One such method, the celebrated $\astar$
algorithm (originally in~\cite{hart68}; also~\cite{pearl}) orders the
search by associating with each node $n$ two values: $g(n)$, the
length of the shortest-path from the initial state to $n$, and
$h(n)$, an estimate of the length of the shortest-path from $n$ to
any goal state (the actual length is $h^*(n)$).  In brief, $\astar$
is an ordered best-first search algorithm, that always examines the
successors of the ``most promising'' node in the search tree, based
on the evaluation function, $f(n) = g(n) + h(n)$.  Thus the number of
nodes expanded is a function of the heuristic function used.

The following definitions will simplify the discussion:

\newenvironment{definition}{\begin{quote}}{\end{quote}}

\begin{definition}
A heuristic function, $h(n)$, is said to be {\em admissible} if
$\forall n [h(n) \leq h^*(n)]$.
\end{definition}

\begin{definition}
A heuristic function, $h(n)$, is said to be {\em monotone} if
$\forall n,n' [f(n) \leq f(n')$ (where $n'$ is a successor of $n$)$]$
(recall that $f(n)$ is determined by $h(n)$).  Monotonicity implies
admissibility~\cite{pearl}.
\end{definition}

$\astar$ terminates when it expands (examines) a goal state for the
first time.  If it uses an admissible heuristic, $\astar$ will have
found the optimal path to this goal state, i.e., $f(n) = g(n) + h(n)
= h^*(I)$.  As it orders the nodes by $f(n) = g(n) + h(n)$, $\astar$
with a monotone heuristic {\em surely expands} all nodes $n$ that
satisfy \(g(n) + h(n) < h^*(I)\), and {\em possibly expands} some of
the nodes that satisfy \(g(n) + h(n) = h^*(I)\), but expands none of
the nodes that satisfy \(g(n) + h(n) > h^*(I)\).  The surely expanded
nodes form a connected component in the state-space graph (containing
the initial state), and those nodes, together with the possibly
expanded nodes, form a larger connected component (containing both
the initial and goal states).

\subsection{Relaxed Models}

As the accuracy of the heuristic function determines the time
complexity (measured in node expansions) of $\astar$, many
researchers have attempted to automate the learning of heuristic
functions.  One proposed method of developing good heuristics is to
``consult simplified models of the problem
domain''~\cite{gaschnig,guida,mandrioli,pearl}.  These simplified
models are generated via {\em constraint-deletion}, i.e., ignoring
selected constraints on the applicability of operators.  Recently,
there has been renewed interest in this and other uses of abstraction
in problem-solving~\cite{knoblock,korf,mostow,tenenberg}, including a
generalization of the notion of simplification~\cite{prieditis} that
will be described below.

Pearl~\cite{pearl} discusses the use of constraint-deletion to
generate three known heuristics for the familiar Eight Puzzle
problem.  He formalizes the problem in terms of domain predicates,
and then describes the single operator in the state-space -- moving a
tile $x$ from position $y$ to position $z$ -- as follows:
\begin{quote}
\begin{tabular}[h]{r l}	
\multicolumn{1}{l}{$Move(x,y,z)$}  \\
preconditions : & $On(x,y) \wedge Clear(z) \wedge
Adj(y,z)$ \\ add list : & $On(x,z) \wedge Clear(y)$ \\ delete list :
& $On(x,y) \wedge Clear(z)$
\end{tabular}  
\end{quote}
Removing preconditions or constraints for this operator creates a
relaxed model of the problem.  Deleting different sets of constraints
yields different relaxed models.

Specifically, when constraints are removed from a problem, new edges
and nodes are introduced into the state-space graph, yielding a
``relaxed'' state-space graph (a supergraph~\cite{gaschnig}).  If the
states and operators of the original problem, $\problem$, are denoted
by the set $S$ and the relation $O \subset S\times S$, then the
relaxed problem, $\problem'$, consists of $S' \supseteq S$ and $O'
\supseteq O$ as well as $I$, and $G$.  Note that relaxed models may
permit certain ``speedup transformations,'' such as factoring into
independent subproblems~\cite{mostow}: our main result underscores
the importance of discovering such optimizations, as discussed in
Section~\ref{interpretation}.

An optimal (i.e., shortest) path between any two given states in
$\problem'$ cannot be longer than the shortest-path between the same
two states in $\problem$, since all paths in $\problem$ are also
paths in $\problem'$.  Thus the length of such a relaxed solution is
a lower bound on the length of an optimal solution to the original
problem, and this information can be used as an admissible heuristic
to speed up a branch-and-bound search algorithm (e.g., $\astar$).
Note that, while our results apply to
``traditional''~\cite{gaschnig,guida,mandrioli,pearl,valtorta80,valtorta84}
relaxed problems obtained by constraint-deletion, other
transformations, more general than constraint-deletion, may be used
to produce admissible heuristics.  Prieditis~\cite{prieditis} gives
conditions for transformations that guarantee the generation of
problems whose solution can be used to compute an admissible
heuristic.  He calls these transformations {\em abstracting} and the
resulting problems {\em abstracted problems}.  Our results do not
necessarily apply to abstracted problems.

Of course, the length of a solution to a further relaxed model
$\problem''$ provides a lower bound on the length of a solution to
$\problem'$.  Thus, if no efficient algorithm\footnote{To be precise,
we describe as {\em efficient} algorithms any techniques that perform
no search in a relaxed model.} is known for solving $\problem'$, we
may use $\problem''$ to provide an admissible heuristic for an
$\astar$ search of $\problem'$.  We may conceive of a search
algorithm (Figure \ref{onealg}) that performs search at many levels
of a relaxed model hierarchy, each level providing heuristic
estimates for the level above.

\vspace{.5in}

\begin{figure}[htb]  
\begin{small}
\begin{verbatim}
           findpath(State I, State G, Problem P) {
               if Efficient_algorithm_for(P) then
                   return(Efficient_algorithm_for(P, I, G))
               else	 
                   return(Astar(I, G, P))
           }
           
           Astar(State I, State G, Problem P) {
           /* a standard A* algorithm, using, for each state s */
           /* h(s) = heur(s,G,P)                               */
           }
           
           heur(State s, State G, Problem P) {
               Problem P' = relax(P);
               return(length(findpath(s,G,P')));
           }
\end{verbatim}
\end{small}
\caption{Search in Relaxed Model Hierarchy}
\label{onealg}
\end{figure}

\newpage

\section{A Motivating Example}

The following example of the application of our result concerns a
heuristic for the Eight Puzzle.  Please refer to the previous section
for a formalization of the puzzle in terms of domain predicates and
operators.  The example is adapted from~\cite{yung}.

In the Eight Puzzle, the tile positions form a bipartite graph of
positions -- each move shifts the blank from one side of the
bipartite graph to the other.  If one colors the puzzle like a
checkerboard and connects adjacent squares by edges, the red squares
will form one side of the bipartite graph and the black squares the
other.  In the Eight Puzzle problem, the blank is constrained to move
to only a small subset of the other side of the graph (i.e., the
adjacent positions).  One may relax this constraint by allowing the
blank to move to any of the positions in the other side of the
bipartite graph.

This Checkerboard Relaxed Adjacency (Check-RA) relaxed model may be
thought of as being somewhere between the original problem and the
Relaxed Adjacency (RA) model~\cite{gaschnig}, because one has deleted
only a part of the adjacency requirement (the RA model results from
deleting the adjacency requirement in moving tiles into the blank
position).  In the Check-RA model, one can think of any given tile
position as being ``adjacent'' to half of the other positions.  To
formalize this, we define two new STRIPS-like predicates (where
$\oplus$ denotes exclusive-or): \begin{quote} \begin{tabular}[h]{r l}
\multicolumn{1}{l}{$RED(y)$} : & $y$ is a red position \\
\multicolumn{1}{l}{$DIFFCOLOR(x,y)$} : & $RED(x) \oplus RED(y)$
\end{tabular} \end{quote} and change the preconditions for MOVE:
\begin{quote} \begin{tabular}[h]{r l}
\multicolumn{1}{l}{$Move(x,y,z)$} \\ preconditions : & $On(x,y)
\wedge Clear(z) \wedge DIFFCOLOR(y,z)$ \\ add list : & $On(x,z)
\wedge Clear(y)$ \\ delete list : & $On(x,y) \wedge Clear(z)$
\end{tabular}  
\end{quote}

In short, if a black position is blank (i.e., clear), only tiles in
red positions may move into it, and {\it vice versa}.  There is no
known algorithm to solve this relaxed model, except for one that
searches the corresponding state space.  However, other relaxed
problems for the Eight Puzzle are more relaxed than Check-RA, e.g.,
Relaxed Adjacency (RA), for which an efficient solution algorithm is
known~\cite{gaschnig,yung}.  We can therefore solve the Check-RA
problem by using $\astar$ with a heuristic computed by solving the RA
problem.  A natural question to ask is whether it is more efficient
to use the heuristic from RA indirectly (to solve Check-RA as just
described) rather than to use it directly to solve the Eight Puzzle.
Clearly, the Check-RA heuristic is never smaller than the RA
heuristic and can therefore prevent $\astar$ from expanding some
nodes that it would expand using the RA heuristic.  On the other
hand, a secondary search procedure (using $\astar$) must be carried
out to compute the Check-RA heuristic.  Is it possible to predict
which one of the two uses of the RA heuristic is the most efficient?
We will answer this question in the affirmative.

\section{The Limitations of Abstraction}

We prove a result that allows us to collapse arbitrary relaxed model
hierarchies to two levels.

Consider that some constraints of a base-level problem $\problem$
have been deleted to yield a relaxed model $\problem'$, whose
solutions provide a heuristic estimate $\hone$ for the base-level
problem.  Unfortunately, there may be no efficient algorithmic
solution to this relaxed model $\problem'$, and to compute $\hone$
one may have to perform a search in $\problem'$ using $\htwo$.  We
will denote this algorithm by $\astaronetwo$.  Consider also the
algorithm $\astartwo$, that uses $\htwo$ directly as a heuristic for
searching $\problem$.  The two algorithms are depicted in Figure
\ref{twoalgs}.

Note that $\astaronetwo$ performs two distinct heuristic searches,
one in $\problem$, and one in $\problem'$.  $\astartwo$ performs only
the search in $\problem$.  We will prove that $\astartwo$ {\em
largely dominates} $\astaronetwo$, i.e., all nodes surely expanded by
$\astartwo$ are also surely expanded by $\astaronetwo$, and all nodes
possibly expanded by $\astartwo$ are also possibly expanded by
$\astaronetwo$. 

\begin{theorem}\label{surely}
If a node $n$ is surely expanded by algorithm $\astartwo$, then node
$n$ is surely expanded by algorithm $\astaronetwo$.

\ 

{\rm 

Assume not.  Consider a node $n$ surely expanded by $\astartwo$, but
not surely expanded by algorithm $\astaronetwo$.  Consider an
ancestor $p$ of $n$ that is the first node on a cheapest path from
$I$ to $n$ that is not surely expanded by $\astaronetwo$:

\begin{equation}\label{eqn0}
g(n) = g(p) + g(p,n)
\end{equation}
\begin{equation}\label{eqn1}
g(p) + \hone(p) \geq h^*(i)
\end{equation}

\ 

$g(p,n)$ denotes the cost of the cheapest path from $p$ to $n$.  By
assumption, computation of $\hone(p)$ does not surely expand $n$:

\begin{equation}\label{eqn2}
g^{\problem'}(p,n) + \htwo(n) \geq \hone(p)
\end{equation}

\ 

$g^{\problem'}(p,n)$ denotes the cost of the cheapest path from $p$
to $n$ in the relaxed model $\problem'$.  Clearly,

\begin{equation}\label{eqn00}
g(p,n) \geq g^{\problem'}(p,n)
\end{equation}

\ 

Substituting inequality \ref{eqn2} into inequality \ref{eqn1}:

\begin{equation}\label{eqn3}
g(p) + g^{\problem'}(p,n) + \htwo(n) \geq h^*(i)
\end{equation}

\ 

and noting from equations \ref{eqn0} and \ref{eqn00} that:

\begin{equation}\label{eqn4}
g(n) \geq g(p) + g^{\problem'}(p,n) \nonumber
\end{equation}

\ 

we derive:

\begin{equation}\label{eqn5}
g(n) + \htwo(n) \geq h^*(i)
\end{equation}

\ 

But by assumption, $n$ is surely expanded by $\astartwo$:

\begin{equation}\label{eqn6}
g(n) + \htwo(n) < h^*(i)
\end{equation}

\ 

which contradicts inequality \ref{eqn5}.  \hfill $\Box$

}

\end{theorem}

\ 

\begin{theorem}\label{possibly}
If a node $n$ is possibly expanded by algorithm $\astartwo$, then
node $n$ is possibly expanded by algorithm $\astaronetwo$.

\ 

{\rm 

Assume not.  Consider a node $n$ possibly expanded by $\astartwo$,
but not possibly expanded by algorithm $\astaronetwo$.  Consider an
ancestor $p$ of $n$ that is the first node on a cheapest path from
$I$ to $n$ that is not possibly expanded by $\astaronetwo$:

\begin{equation}\label{peqn0}
g(n) = g(p) + g(p,n)
\end{equation}
\begin{equation}\label{peqn00}
g(p,n) \geq g^{\problem'}(p,n)
\end{equation}
\begin{equation}\label{peqn1}
g(p) + \hone(p) > h^*(i)
\end{equation}

\ 

By assumption, computation of $\hone(p)$ does not possibly expand $n$:

\begin{equation}\label{peqn2}
g^{\problem'}(p,n) + \htwo(n) > \hone(p)
\end{equation}

\ 

Substituting inequality \ref{peqn2} into inequality \ref{peqn1}:

\begin{equation}\label{peqn3}
g(p) + g^{\problem'}(p,n) + \htwo(n) > h^*(i)
\end{equation}

\ 

and noting from equations \ref{peqn0} and \ref{peqn00} that:

\begin{equation}\label{peqn4}
g(n) \geq g(p) + g^{\problem'}(p,n) \nonumber
\end{equation}

\ 

we derive:

\begin{equation}\label{peqn5}
g(n) + \htwo(n) > h^*(i)
\end{equation}

\ 

But by assumption, $n$ is possibly expanded by $\astartwo$:

\begin{equation}\label{peqn6}
g(n) + \htwo(n) \leq h^*(i)
\end{equation}

\ 

which contradicts inequality \ref{peqn5}.  \hfill $\Box$

}

\end{theorem}

\ 

\section{Interpretation of Results}
\label{interpretation}

As anticipated at the beginning of the previous section, Theorems 1
and 2 imply that $\astartwo$ largely dominates $\astaronetwo$.  This
is precisely the sense in which $\astar$ is said to be {\em optimal}
~\cite[p. 85]{pearl}: given a monotone heuristic $h$, $\astar$ {\em
largely dominates} every admissible algorithm that uses $h$.  The
relation misses some important distinctions.  For example, greedy
algorithms, that may operate in polynomial time, can be viewed as
heuristic searches with optimal tie-breaking rules to choose among
the possibly expanded nodes.  Large domination does not imply that
all nodes expanded by $\astartwo$ will necessarily be expanded by
$\astaronetwo$.  In fact there are some tie-breaking rules (for
ordering expansion of nodes with equal $f$ values) for which
$\astaronetwo$ will expand fewer nodes than $\astartwo$.
Traditionally, however, large domination is equated with superior
efficiency, as the number of nodes $n$ for which $g(n) + h(n) =
h^*(i)$ is assumed to be small: for example, if either $g(n)$ or
$h(n)$ are continuous valued functions~\cite[p. 85]{pearl} or $h$
is non-pathological (as defined in~\cite[p. 522]{pearl85}), such
ties are rare or non-existent.

The theorems extend results of Valtorta~\cite{valtorta84}, who proved
that brute-force, uniform-cost search (Dijkstra's algorithm) is never
less efficient than a heuristic search that relies on uniform-cost
search of relaxed models for the computation of heuristics.  This
follows from our results by considering the special case in which
$\htwo$ is uniformly zero, yielding uniform-cost search.  However,
Valtorta's original result is for a sharper notion of ``more
efficient than'': the number of node expansions, rather than large
domination.  The reason for this discrepancy lies in the fact that
Dijkstra's algorithm expands all and only the nodes for which
$g^{*}(n) < h^{*}(i)$, while $\astar$ with monotone heuristics may
expand some nodes for which $g^{*}(n) + h(n) = h^{*}(i)$: for
Dijkstra's algorithm, the set of possibly expanded nodes coincides
with the set of surely expanded nodes.  Analogously, while Dijkstra's
algorithm is optimal with respect to the measure ``number of node
expansions'' within the class of blind forward unidirectional
algorithms~\cite{valtorta84}, $\astar$ is optimal in that it largely
dominates all informed algorithms that use consistent heuristics.
``Large domination'' is a weaker form of optimality than ``fewer node
expansions,'' as is explained by Dechter and Pearl in~\cite{pearl83},
and especially in~\cite{pearl88}.

We observe that $\astar$ with monotone heuristics, just like
Dijkstra's algorithm, never expands the same node twice.  Since all
heuristics computed by solving a relaxed problem are
monotone~\cite{pearl,valtorta84}, we can use ``the number of expanded
nodes'' and ``the number of node expansions'' interchangeably.
Therefore our results hold for the cost measure ``number of node
expansions,'' as well as for the cost measure ``number of expanded
nodes.''

While Valtorta~\cite{valtorta84} showed that there is no advantage to
using $\astar$ in certain situations, we show that, in certain
situations, there is no advantage to using a relaxed-model hierarchy
within an $\astar$ algorithm.  Theorems 1 and 2 indicate that in
reducing the cost of finding optimal solutions, a heuristic is
effective only when it is computable by an efficient algorithm -- the
cost of computing a heuristic function by a secondary search exceeds
the savings offered by the heuristic.  This possibility was
recognized in the design of ABSOLVER~\cite{mostow}, where ``speedup
transformations'' are used, as well as in other recent
work~\cite{kibler}.  Our result shows (in a precise and general way)
that speedup transformations are not only useful, but necessary.

Some readers may object to our use of the word ``search'' in the
preceding paragraph, since, after all, a search in a much reduced
state space is sufficient to turn an inefficient procedure into a
practical, efficient one.  In the spirit of~\cite{guida,valtorta84}
we use the word search to signify that a relaxed problem is solved by
search in {\em its} state space (called {\em underlying graph}
in~\cite{guida} and {\em skeleton} in~\cite{valtorta84}) and view the
construction of a reduced state space as an instance of a ``speedup
transformation.''  A way of reducing the state space for a (relaxed)
problem is to factor (or decompose) it into independent subproblems.
Decomposability is possible when the goal state of a state-space
search problem can be achieved by solving its subgoals independently.
As discussed in~\cite{mostow,pearl}, the state-space for each
independent subproblem can be so much smaller than the original
state-space that computation of the heuristic by searching the
factored problem is computationally advantageous over blind search.
We illustrate these considerations with an example adapted
from~\cite{mostow}.

\begin{figure}[htbp]
\begin{center}
\epsfysize=3in\leavevmode\epsfbox{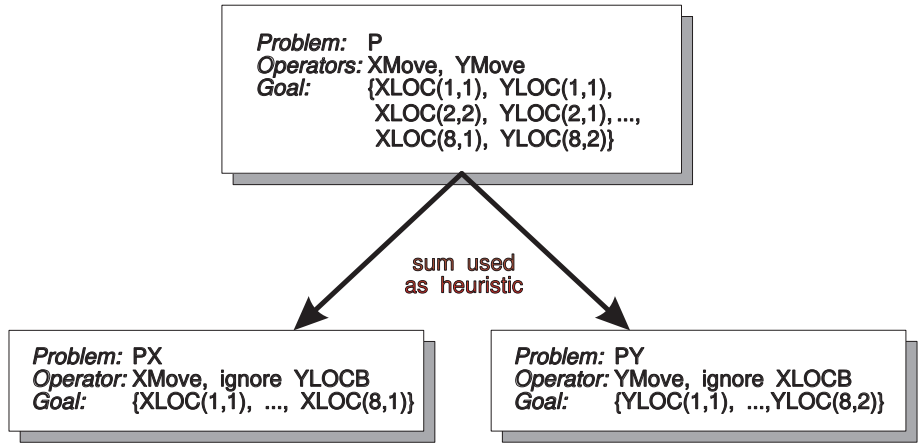}
\end{center}
\caption{Two independent subproblems to compute the X-Y heuristic}
\label{XY}
\end{figure}

Mostow and Prieditis describe a heuristic (called X-Y) that is never
smaller than the well-known Manhattan Distance (MD) heuristic.  It
corresponds to a relaxed problem in which ``a horizontal move is
allowed only into a column containing the blank, and a vertical move
is allowed only into the row containing the blank.  X-Y is therefore
more accurate than Manhattan Distance, which ignores the blank
completely.''  This relaxed problem can be obtained from a
representation of the tiles in the Cartesian plane\footnote{We use 
the traditional goal state with the blank in the middle and tiles
arranged clockwise around the border, starting with tile 1 in the top left
corner.
Mandrioli
et al.~\cite{mandrioli} used the Cartesian representation for the
Eight Puzzle in the first published examples of relaxed problems, but
they did not provide the X-Y heuristic as an example.}, by dropping
all information about the position of the blank in the Y coordinate
(YLOCB) from moves in the X coordinate (XMove) and all information
about the position of the blank in the X coordinate (XLOCB) from
moves in the Y coordinate (YMove).  The X-Y relaxed problem can be
decomposed into two subproblems (or {\em factors}), corresponding to
each coordinate.  The goal of the first subproblem is to place each
tile in its correct column, while the goal of the second subproblem
is to place each tile in its correct row.  Since we dropped from the
list of preconditions of the move operator in a coordinate all
information about the position of the blank in the other coordinate,
we obtain two independent subproblems, as described in
Figure~\ref{XY}.

No algorithm is known to solve either of the independent subproblems
without searching the corresponding state-space.  However, since MD
is a relaxed problem of X-Y for which a very efficient solution
algorithm is known, it is possible to solve X-Y by using $\astar$
with a heuristic computed by solving MD.  Now consider the cost of
computing X-Y as just outlined.  This computation involves search.
However, the search is in the reduced state-spaces of two independent
subproblems of the X-Y relaxed problem, and therefore the results
described in this paper are not applicable.  In other words, 
the theorems we have proved in this paper are not strong enough to predict
when a heuristic derived from relaxation and factoring is cost effective,
and it is
{\em not} possible to conclude whether the computation of the X-Y
heuristic is cost effective simply on the basis of our results.  In
fact, it has been determined empirically that computation of the X-Y
heuristic as described is not cost effective.  Mostow and Prieditis
report that ``unfortunately, even with such guidance [i.e., the
heuristic computed by solving MD] the overall search time turns out
to be about six times slower than using Manhattan distance alone.''
In other words, direct use of MD is more efficient than indirect use
of MD to compute X-Y.  This shows that heuristics computed by
searching factors of relaxed problems need not be cost effective.  On
the other hand, the Box Distance heuristic of the Rooms World
problem, when computed by search of factors of a
relaxed problem, is cost effective~\cite{prieditis}.
This
shows that heuristics computed by searching factors of relaxed
problems can be cost effective.
Additional work is needed to find conditions under which heuristics
computed by searching factors of relaxed problems are 
cost effective, and conditions under which they are not.

In conclusion, we have shown that relaxed model hierarchies (chains
of successively more abstract problems) collapse to only two levels
-- the base-level, and the least-relaxed model for which algorithmic
solutions are known. Exploiting the power of abstraction will require
either the development of techniques for synthesizing efficient
algorithms from relaxed-model problem-descriptions, or the
development of new problem relaxation transformations that are not
subject to the fundamental limitations of the theorems presented
here~\cite{kibler,mostow,prieditis}.  The most promising example of
these approaches is the ABSOLVER system~\cite{mostow,prieditis}, which
exploits subproblem factoring transformations to reduce the cost of
search in relaxation hierarchies.

\section{Acknowledgements} 
We would like to thank two anonymous referees for their insightful
and detailed comments on earlier drafts of this paper and Armand
Prieditis for several interesting conversations.  Marco Valtorta
thanks Marco Somalvico and Giovanni Guida for introducing him to
research on auxiliary problems and Jack Mostow for a conversation in
which he shared some (at the time) unpublished details of the
ABSOLVER system.

\end{document}